THE BAYESIAN APPROACH TO CONTINUAL LEARNING: AN OVERVIEW

# The Bayesian Approach to Continual Learning: An Overview

**Tameem Adel** TAMEEM.HESHAM@GMAIL.COM
*NPL, Maxwell Centre, University of Cambridge,*
*JJ Thomson Avenue, Cambridge, CB3 0HE, United Kingdom.*

## Abstract

Continual learning is an online paradigm where a learner continually accumulates knowledge from different tasks encountered over sequential time steps. Importantly, the learner is required to extend and update its knowledge without forgetting about the learning experience acquired from the past, and while avoiding the need to retrain from scratch. Given its sequential nature and its resemblance to the way humans think, continual learning offers an opportunity to address several challenges which currently stand in the way of widening the range of applicability of deep models to further real-world problems. The continual need to update the learner with data arriving sequentially strikes inherent congruence between continual learning and Bayesian inference which provides a principal platform to keep updating the prior beliefs of a model given new data, without completely forgetting the knowledge acquired from the old data. This survey inspects different settings of Bayesian continual learning, namely task-incremental learning and class-incremental learning. We begin by discussing definitions of continual learning along with its Bayesian setting, as well as the links with related fields, such as domain adaptation, transfer learning and meta-learning. Afterwards, we introduce a taxonomy offering a comprehensive categorization of algorithms belonging to the Bayesian continual learning paradigm. Meanwhile, we analyze the state-of-the-art while zooming in on some of the most prominent Bayesian continual learning algorithms to date. Furthermore, we shed some light on links between continual learning and developmental psychology, and correspondingly introduce analogies between both fields. We follow that with a discussion of current challenges, and finally conclude with potential areas for future research on Bayesian continual learning.

## 1. Introduction

Continual learning (also referred to as incremental learning or lifelong learning) is an online paradigm where (non i.i.d.) data continuously arrive. The data distribution can potentially change over time (Schlimmer & Fisher, 1986; Sutton & Whitehead, 1993; Ring, 1995, 1997; Kirkpatrick et al., 2017; Lee et al., 2017; Shin et al., 2017; Schmidhuber, 2018; Ahn et al., 2019; Riemer et al., 2019; Buzzega et al., 2020; Liu et al., 2020; Mirzadeh et al., 2020; Yoon et al., 2020; Beaulieu et al., 2021; Mundt et al., 2022; Romero et al., 2022; Wu et al., 2022). Upon learning from upcoming data, the continual learner should not forget the knowledge acquired from previous data, a phenomenon referred to as catastrophic forgetting (McCloskey & Cohen, 1989; Ratcliff, 1990; Robins, 1993, 1995; French, 1999; Pape et al., 2011; Srivastava et al., 2013; Achille et al., 2018; Diaz-Rodriguez et al., 2018; Kemker et al., 2018; Zeno et al., 2018; Parisi et al., 2019; Pfulb & Gepperth, 2019; Ebrahimi et al., 2020; Gupta et al., 2020; Banayeeanzade et al., 2021; Ke et al., 2021; Ostapenko et al., 2021; Wang et al., 2021; Karakida & Akaho, 2022; Lin et al., 2022; Miao et al., 2022). The continual learner should as well adapt to the distributional shift occurring across the sequential learning phases. A balance must therefore be achieved in continual learning (CL) between stability, to preserve previous knowledge, and adaptation. This is commonly referred to in the CL literature as the stability-plasticity tradeoff (Kim et al., 2023; Adel, 2024, 2025). Stability in a CL context refers to the ability of a model to retain existing knowledge, whereas plasticity signifies the capacity of the model to adapt and learn new information. Updates must be performed incrementally where the data available at each stage comprise solely the new data. Due to privacy, security and computational constraints, access to the old data is prohibited (Adel et al., 2020; Smith et al., 2023).

The capability of humans to learn from the past, even given limited experience, is superior to that of the machines (Taylor & Stone, 2009; Chen & Liu, 2016; Finn et al., 2017; Li et al., 2018; Rostami et al., 2020). This is partly due to the fact that humans possess a mechanism which is adept at gaining and tuning knowledge throughout their different life stages (Li et al., 2018; Parisi et al., 2019). In contrast, until recently, machine learning models, in particular neural networks, have solely addressed i.i.d. data, which results in challenges, e.g. the aforementioned phenomenon of catastrophic forgetting, upon encountering sequentially arriving, non-stationary data.

As such, CL frameworks should be capable of preserving previous knowledge, without storing previous data, and of updating such knowledge according to new data. This strikes a substantial analogy with the Bayesian inference paradigm where a probability distribution over the parameters of the model expresses the current status of knowledge given the data observed so far. Upon the arrival of new data, the aforementioned status can then be thought of as the prior which, along with the knowledge deduced from the new data (the likelihood), results in inferring the posterior (Nguyen et al., 2018). This posterior can in turn be thought of as the new prior upon the arrival of further data in the future, and so on. Several previous CL frameworks have based their reasoning on this congruence between CL and the Bayesian paradigm.

We kick off this survey by defining continual learning (CL) as well as Bayesian continual learning (BCL), and also formalize the CL settings which will be covered herein, namely task-incremental learning and class-incremental learning (Section 2). We then proceed by clarifying the distinction with related fields, like meta-learning and domain adaptation (Section 3). We move on to present a taxonomy of BCL (Section 4). In addition, we provide an outline of the most seminal BCL algorithms to date. We then establish a few associations between BCL and developmental psychology (Section 5). We finally conclude with a summarization of our view on the major challenges that directly influence BCL algorithms, like catastrophic forgetting, and also propose a few areas that we believe are potentially promising for future research on BCL (Section 6).

## 2. Formulation

The most widely used CL settings are task-incremental learning and class-incremental learning.

In task-incremental learning, the training procedure is divided into separate phases where each phase corresponds to a task[1]. The main assumption in (task-aware) task-incremental learning is that task identities are observed during both the training and test procedures.

A typical task-incremental learner encounters a set of $m$ sequentially arriving datasets, $\mathcal{D}_t = \{\boldsymbol{x}_t^n, \boldsymbol{y}_t^n\}_{n=1}^{N_t}$, where $t \in \{1, 2, \ldots, m\}$ is the task index, and $N_t$ is the size of the training dataset of task $t$. Data points are depicted by input features $\boldsymbol{x} \in \mathcal{X}$ and corresponding labels $\boldsymbol{y} \in \mathcal{L}_t$, where $\mathcal{X}$ is the input space. For classification[2], the label space for task $t$, $\mathcal{L}_t$, is a finite set consisting of a set of classes (i.e. labels). Typically speaking, in both task-incremental and class-incremental learning,

1. A task can as well be referred to as a "session".
2. Similar to the majority of works in the area, we focus on classification herein, but the extension to label spaces like regression is straightforward.

solely the training dataset $\mathcal{D}_t$ is available during the training procedure of the corresponding task $t$ (Figure 1). Due to privacy constraints, data belonging to the previous tasks are not available. For each test data point $\boldsymbol{x}_t$, the task-incremental learner mainly aims to predict a class label $\boldsymbol{y}_t|\boldsymbol{x}_t$, where $\boldsymbol{y}_t \in \mathcal{L}_t$, and both the task $t$ and, consequently, the respective label space $\mathcal{L}_t$ are observed by the learner a priori. In other words, the pool of potential labels from which the class label $\boldsymbol{y}_t$ can be selected during the test (inference) procedure in task-incremental learning is exclusively depicted by the labels belonging to the respective task $t$ (Figure 2).

On the other hand, in class-incremental learning (CIL), $m$ datasets, $\mathcal{D}_t = \{\boldsymbol{x}_t^n, \boldsymbol{y}_t^n\}_{n=1}^{N_t}$, are also sequentially presented to the class-incremental learner where $N_t$ is the size of dataset $t$, $t \in \{1, 2, \ldots, m\}$, is the task index. Note that, in CIL, the sets of classes belonging to different tasks are disjoint, that is $\forall i, j \in \{1, 2, \ldots, m\}$, $i$ and $j$ are task indices, and $i \neq j$, $\mathcal{L}_i \cap \mathcal{L}_j = \varnothing$. In contrast to the task-incremental learning setting, a class-incremental learner aims to predict a class label $\boldsymbol{y}$ for each test data point $\boldsymbol{x}$ where the label $\boldsymbol{y}$ can belong to *any* previously encountered label space, irrespective of the task identity $t$, since there is no task identity information available at the disposal of the class-incremental learner during the test (inference) procedure. Assuming that the class-incremental learner has already encountered $m$ tasks, the predicted class label $\boldsymbol{y}$ can therefore belong to any of the label spaces of the previously encountered tasks, $\boldsymbol{y} \in \bigcup_{j=1}^{m} \mathcal{L}_j$. For example, think of a handwritten digit recognition model where it is required to identify the characteristics of each handwritten digit for prediction purposes. Assume that the class-incremental learner encounters the problem of classifying digits "1" vs. "2" as the first task. Given their rather different characteristics, it might be relatively uncomplicated for the class-incremental learner to recognize all the "1" digits during the first task. Assume then that the second (or, more generally, a successive) task that the class-incremental learner encounters is to classify digits "7" vs. "8". At this point, the mission of the class-incremental learner becomes much more challenging since it is now required to simultaneously distinguish between all of the four handwritten digits during inference, which involves the tricky problem of differentiating between handwritten versions of the digits "1" and "7" (Figure 2). In contrast, a task-incremental learner which has encountered the same tasks would have never needed to differentiate between the handwritten digits "1" and "7", neither during training nor during inference, since they belong to different tasks, and since the task identity is always observed (i.e. given as input) during training as well as during the test (inference) procedure.

Compared to the standard CIL setting outlined above, few-shot class-incremental learning (FSCIL) induces an additional level of difficulty given the scarcity of data available for every class appearing after the first task. The first task in FSCIL ($t = 1$) is commonly referred to as the base task, whereas the successive tasks (i.e. from the second task onwards, $t \geq 2$) are referred to as few-shot tasks. For the few-shot tasks, solely a scarce amount of data is available for every class (Rebuffi et al., 2017; Gidaris & Komodakis, 2018; Tao et al., 2020; Achituve et al., 2021; Ahmad et al., 2022; Peng et al., 2022; Song et al., 2023; Wang et al., 2023b; Zhou et al., 2023; Zhao et al., 2024), whereas the base dataset $\mathcal{D}_1$ is a large-scale training dataset consisting of multiple base classes, with relatively abundant data per class (compared to the subsequent, few-shot tasks, $t > 1$). In other words, for any few-shot task, $t > 1$: $N_1 >> N_t$. Furthermore, for few-shot tasks, $t > 1$, the amount of available data per class in the FSCIL setting is also more limited than the corresponding tasks in a vanilla CIL setting. For the datasets of the incremental few-shot tasks, $\mathcal{D}_t, t > 1$, the $C$-way $K$-shot FSCIL setting refers to the fact that the respective task comprises $C$ classes and $K$ training data points per class. This means that, for a few-shot task $t, t > 1$, the total size of the training dataset is $N_t = C \times K$.

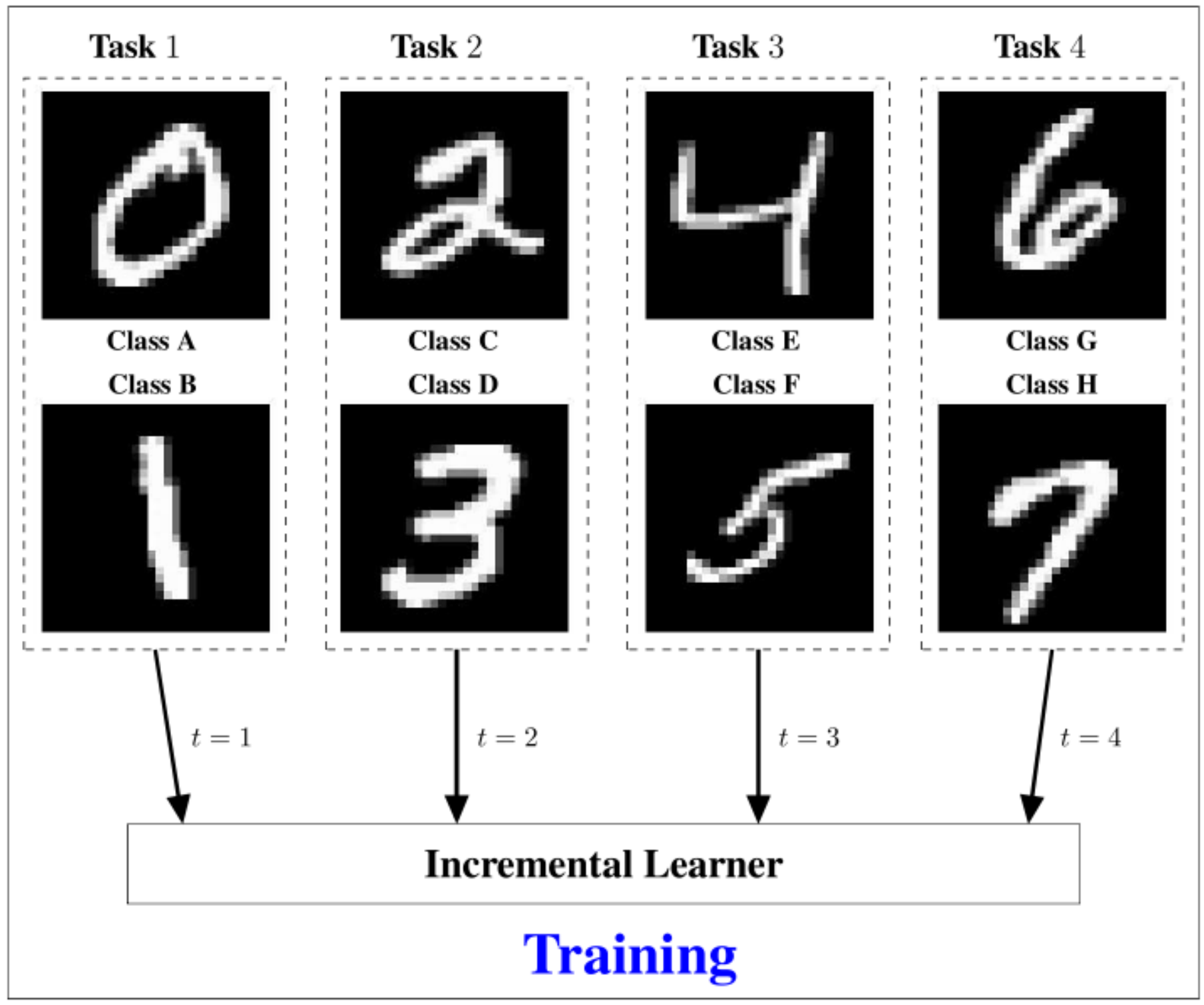

Figure 1: The training phase for an incremental learner. In both task-incremental learning (TIL) and class-incremental learning (CIL), the learner can access the task identity of each class during *training*.

Refer to the label prediction loss for task $t$ as $L^t(\cdot,\cdot)$. The principal goal in continual learning is to minimize the overall expected loss: $\sum_{t=1}^{m} \mathbb{E}_{(\boldsymbol{x}_t^n,\boldsymbol{y}_t^n)_{n=1}^{N_t}} \left[L^t(f_t(\boldsymbol{x}_t^n;\boldsymbol{\theta}_t),\boldsymbol{y}_t^n)\right]$, where $f_t$ depicts the label predictor of task $t$, and $\boldsymbol{\theta}_t$ represents the parameters of the respective label predictor, i.e. the parameters of the CL model that is learned to ultimately predict test points from every CL task encountered so far.

### 2.1 Evaluation Metrics

We shed light here on the most widely used evaluation metrics in CL.

During the test procedure, the continual learner has access to test data belonging to all the tasks encountered so far. Assume that the continual learner has thus far encountered $m$ tasks, the corresponding performance evaluation will then include a test set from every task $t, t \in \{1, 2, \dots, m\}$. Let $\mathcal{A}_{i,j}$ denote the test classification accuracy of the continual learner on task $t = j$ after having learned from (i.e. after having performed the training procedure on) task $t = i$. In addition, let $\mathcal{A}_j$ denote the test classification accuracy on task $t = j$ after randomly initializing a reference model and then learning solely from data belonging to the same task $t = j$ (Lopez-Paz & Ranzato, 2017a). The overall average accuracy of a continual learner after learning from $m$ tasks can accordingly be defined as follows:

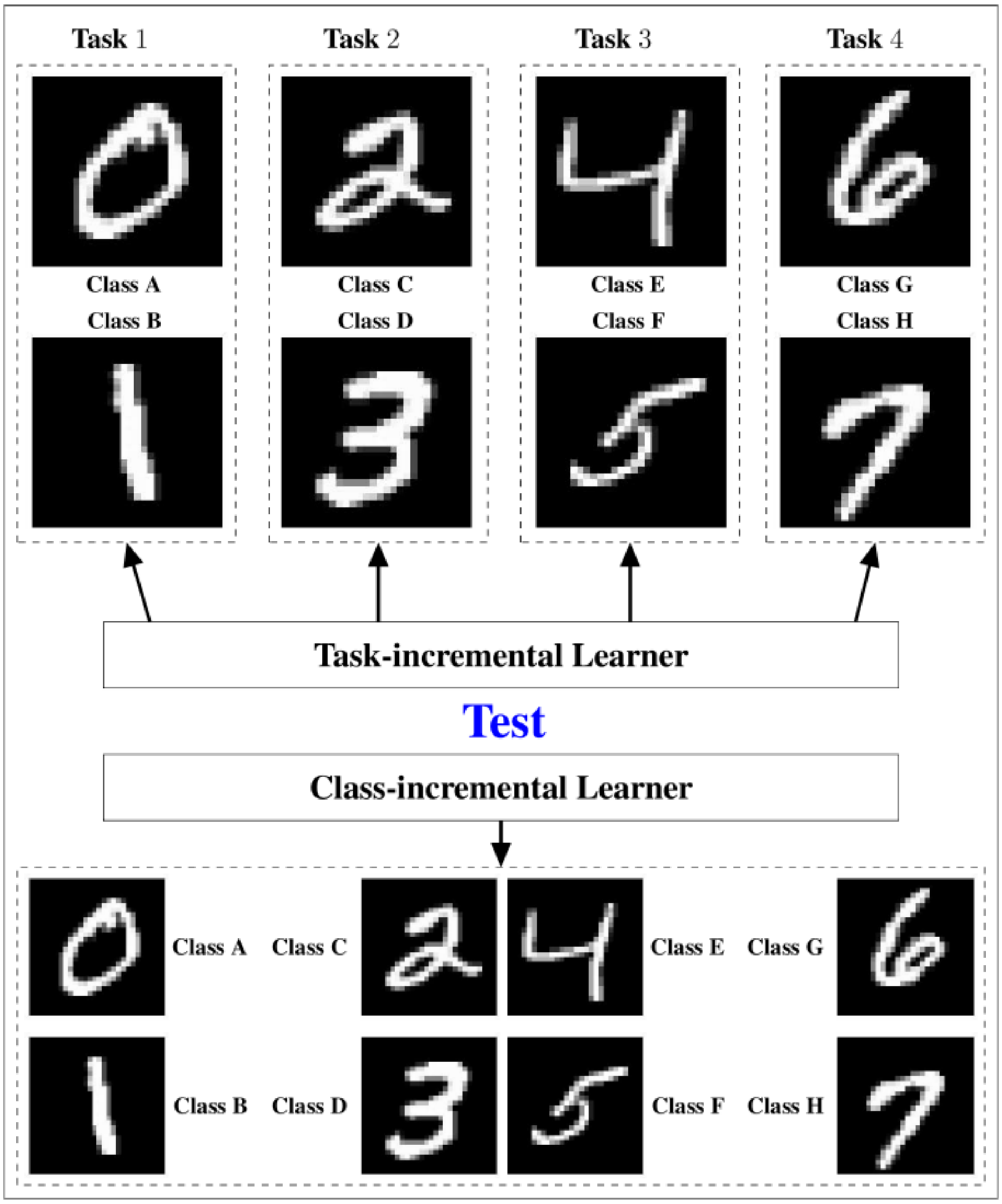

Figure 2: The test phases of a task-incremental learner and of a class-incremental learner. Task IDs of every class are available for the task-incremental learner during the test phase (upper part of the figure). In contrast, a class-incremental learner must address the tricky issue of distinguishing between all the previously encountered classes (regardless of the task they originally belonged to) since the task IDs are *not* available for the class-incremental learner during the test phase (lower part of the figure). As mentioned in the text, the class-incremental learner will then have to distinguish between digits "1", "2", . . ., "7" and "8" (including the onerous problem of distinguishing between digits "1" and "7") without ever having had the opportunity to train on all of such digits simultaneously.

$$\mathrm{AA}_m = \frac{1}{m}\sum_{j=1}^{m}\mathcal{A}_{m,j}. \tag{1}$$

The larger the value of the overall average accuracy after $m$ tasks, $\text{AA}_m$, the better. It is also of paramount importance to evaluate catastrophic forgetting. Backward transfer (BWT) is one of the main task-incremental learning metrics used for such a purpose. The BWT metric basically functions by assessing the impact that learning a task $t = i$ imposes on the learning performance of a previous task $t = j$, where $j < i$ (Figure 3). Again, a large positive BWT value is desirable since it denotes that learning the subsequent task $t = i$ has resulted in an improved performance on the previous task $t = j$, whereas the other extreme, which entails a high degree of catastrophic forgetting (massively degraded performance on the previous task $t = j$), holds with a large negative value of the BWT metric. The backward transfer (BWT) metric is defined as follows:

$$BWT = \frac{1}{m-1} \sum_{j=1}^{m-1} (\mathcal{A}_{m,j} - \mathcal{A}_{j,j}). \tag{2}$$

Note that it is pointless to try and assess the BWT metric on the final task $t = m$. Another important metric, which is referred to as forward transfer (FWT), measures the impact that learning a task $t = j$ imposes on the performance of a future task $t = i$, $i > j$ (Figure 3). The larger the FWT metric value, the better. We adopt the most widely used definition of FWT (Lopez-Paz & Ranzato, 2017b) which compares the impact of learning task $t = j - 1$ on the future task $t = j$, with the performance on task $t = j$ after random initialization, $\mathcal{A}_j$:

$$FWT = \frac{1}{m-1} \sum_{j=2}^{m} (\mathcal{A}_{j-1,j} - \mathcal{A}_j). \tag{3}$$

The indices of the summation in (3) begin at task $t = 2$ since it is senseless to evaluate the forward transfer (FWT) metric on the first task. Larger FWT values denote better performance.

The overall average accuracy defined in (1) is also adopted for CIL and FSCIL to evaluate the overall classification accuracy over all the classes encountered so far. Due to the massive importance of monitoring the evolution of the classification accuracy throughout the progression of the consecutive incremental tasks, and not only after the final task, in the CIL setting, another average accuracy metric is adopted for CIL and FSCIL (Wang et al., 2023a; Zhou et al., 2023) in order to enable the tracking of the historical variation of the average accuracy. This is referred to as the average incremental accuracy, $\overline{\text{AIA}}$, which is computed as follows:

$$\overline{\text{AIA}} = \frac{1}{m} \sum_{t=1}^{m} \text{AA}_t \tag{4}$$

The most commonly used FSCIL metric to estimate catastrophic forgetting is referred to as the performance dropping rate (PD). The PD metric (Zhang et al., 2021) is defined as the difference between the overall average accuracy after the base task/session, $\text{AA}_1$, and the overall average accuracy after the final few-shot task, $\text{AA}_m$, assuming that there is a total of $m$ tasks available at the disposal of the few-shot class-incremental learner:

$$\text{PD} = \text{AA}_1 - \text{AA}_m, \tag{5}$$

where $\mathrm{AA}_1$ denotes the classification accuracy after the base session, and $\mathrm{AA}_m$ refers to the overall average classification accuracy after learning from the final task $t = m$. Unlike all the other CL metrics defined herein, lower values of the PD metric signify better performance levels, since this means that the respective few-shot class-incremental learner has managed to be less forgetful of the first task, and has thus achieved the desirable outcome of higher retention levels.

Figure 3: A schematic figure denoting backward transfer (BWT) and forward transfer (FWT) in CL. The main aim of BWT is to assess the impact of learning a task $t$ on the performance of a previous task $t-1$. In contrast, FWT estimates the impact that learning a task $t$ imposes on the performance of a future task $t+1$. Note that we display $t-1$ and $t+1$ as solely one example of a previous task and a future task, respectively, for the sake of presentation clarity.

### 2.2 Continual Learning via Bayesian Inference

Refer to a discriminative continual learner which outputs probabilistic predictions in the form of a probability distribution $p(\boldsymbol{y}|\boldsymbol{\theta}, \boldsymbol{x})$ over the output classes $\boldsymbol{y}$ given input features $\boldsymbol{x}$ and model parameters $\boldsymbol{\theta}$. The parameters $\boldsymbol{\theta}$ of this continual learner are learned from the aforementioned set of $m$ sequentially arriving datasets, $\mathcal{D}_t = \{\boldsymbol{x}_t^n, \boldsymbol{y}_t^n\}_{n=1}^{N_t}$, $t \in \{1, 2, \ldots, m\}$ is the task index, and $N_t$ is the size of the training dataset of task $t$. A typical Bayesian approach to CL would begin by defining a prior distribution $p(\boldsymbol{\theta})$ over the model parameters $\boldsymbol{\theta}$ (Nguyen et al., 2018; Kessler et al., 2023). After having encountered $m$ tasks, the posterior is then computed based on Bayes' rule:

$$p(\boldsymbol{\theta}|\mathcal{D}_{1:m}) \propto p(\boldsymbol{\theta}) \prod_{t=1}^{m} p(\mathcal{D}_t|\boldsymbol{\theta}) = p(\boldsymbol{\theta}) \prod_{t=1}^{m} \prod_{n=1}^{N_t} p(\boldsymbol{y}_t^n|\boldsymbol{\theta}, \boldsymbol{x}_t^n) \tag{6}$$

Note that what embodies a prior and a posterior varies at each learning phase. A Bayesian continual learner which is about to learn from task $t = m$, based on the respective dataset $\mathcal{D}_m$, takes the previous posterior (i.e. what was formerly considered the posterior at the previous task $t = m-1$) as its own prior, multiplies it with the likelihood, and then normalizes the outcome according to Bayes' rule:

$$p(\boldsymbol{\theta}|\mathcal{D}_{1:m}) \propto p(\boldsymbol{\theta}|\mathcal{D}_{1:m-1})p(\mathcal{D}_m|\boldsymbol{\theta}) \tag{7}$$

There are different manners via which the posterior over CL model parameters, $p(\boldsymbol{\theta}|\mathcal{D}_{1:m})$, can be learned, as will be detailed later on. Ultimately, given a test data point $\boldsymbol{x}^*$, prediction (of the corresponding $\boldsymbol{y}^*$) is performed based on the predictive posterior which can be denoted as follows:

$$p(\boldsymbol{y}^*|\boldsymbol{x}^*, \mathcal{D}_{1:m}) = \int p(\boldsymbol{\theta}|\mathcal{D}_{1:m})p(\boldsymbol{y}^*|\boldsymbol{\theta}, \boldsymbol{x}^*)\, d\boldsymbol{\theta}, \tag{8}$$

where the term $p(\boldsymbol{y}^*|\boldsymbol{\theta}, \boldsymbol{x}^*)$ depicts the label predictor with parameter values $\boldsymbol{\theta}$.

Bayesian continual learning entails several advantages since it offers a principled approach to retain previous knowledge, as well as adapt to the newly encountered knowledge, via quantifying uncertainty over the model parameters.

## 3. Related Paradigms

Given the fact that there are machine learning paradigms which may outwardly seem similar to continual learning, it is important to highlight the nuances between the characteristics of such paradigms and continual learning. In this section, we provide brief descriptions of these closely related paradigms, and summarize the main differences with continual learning (Figure 4).

**Transfer learning (TL).** A TL model is trained on a source task $t_{src}$ so that it can exploit this knowledge to perform on a different, yet related, task. The latter is referred to as the target task $t_{trgt}$. Note that the fact that these are two different tasks means that the label spaces of the source and target tasks can as well be different, i.e. $\mathcal{L}_{src} \neq \mathcal{L}_{trgt}$. Since the TL performance is assessed solely based on the target task, TL models do not take the forgetting aspect into consideration; they solely focus on performing forward transfer of the knowledge acquired from one source task to one target task.

**Domain adaptation (DA).** Similar to TL, catastrophic forgetting is not considered during the optimization of DA models. A typical DA setting involves a source domain and a target domain. DA models are trained on the source domain in order to adapt this knowledge to the (similar yet not identical) target domain. Performance of a DA model is solely assessed on the target domain. The training data in the target domain consist of either unlabelled data only or unlabelled data along with few labelled data examples. DA can be viewed as a special case of TL, since in DA there are source and target domains (and not tasks). In other words, the source and target environments can solely differ in their input and output distributions. The most widely used DA setting is covariate shift (Adel & Wong, 2015; Adel et al., 2017) where (solely) the input distributions are different $p(X_{src}) \neq p(X_{trgt})$. In the more general setting of DA, the joint input and output distributions can differ among the source and target domains $p(X_{src}, Y_{src}) \neq p(X_{trgt}, Y_{trgt})$.

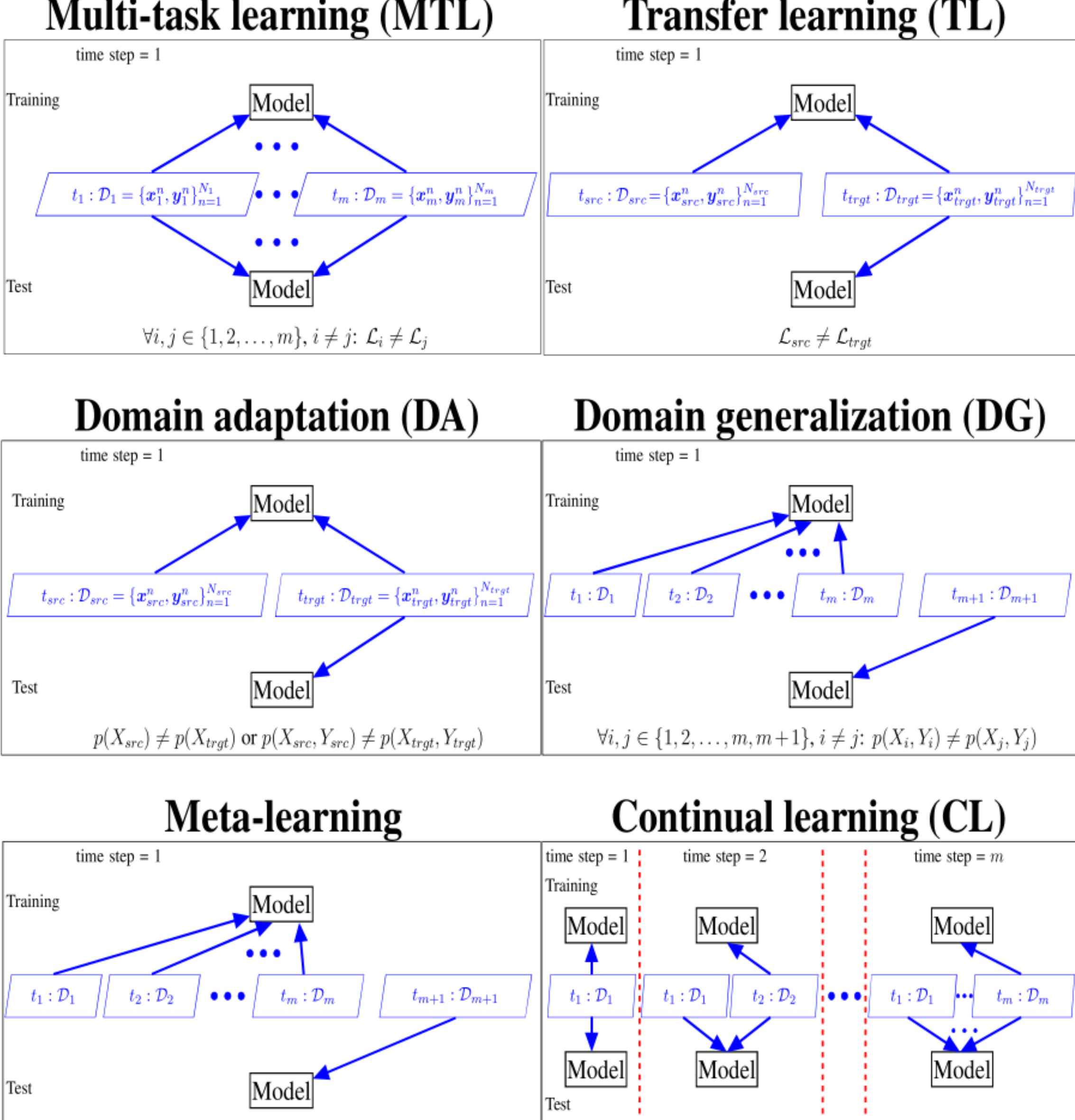

Figure 4: The main characteristics of machine learning paradigms which are related to continual learning (CL), and how they differ from CL. One of the most distinguishing features of CL is the prevalence of the catastrophic forgetting aspect. This is mainly due to the fact that tasks arrive sequentially, along with the manner based on which performance evaluation is performed in CL where all of the already encountered tasks (not only the latest task) are taken into consideration during the test procedure. More details are provided in Section 3.

**Domain generalization (DG).** In DG, the learner encounters several different, yet related, domains as the input. The goal is to learn how to generalize to an unobserved test domain. Domain generalization is also referred to as out-of-distribution generalization. Similar to DA, in DG the i.i.d. assumption that both the training and test data are identically and independently distributed does not hold. Unlike DA, a DG learner cannot access any data (labelled or unlabelled) nor information about the test domain during its training phase. Similar to DA and TL, DG is only focused on a forward transfer perspective, with no consideration whatsoever to knowledge retention or forgetting, since it is only assessed based on the test domain.

**Multi-task learning (MTL).** The training procedure of an MTL model is based on a set of several related (yet not identical) training tasks. The main objective of a multi-task learner is to jointly learn from the training tasks simultaneously in order to optimize the resulting model for each training task. In other words, MTL models do not aim to generalize to any other (unobserved) tasks; they are both trained and tested on the same tasks.

**Meta-learning.** It is also referred to as learning to learn since the learner is presented with a large number of different, yet related, learning tasks, and the main objective is to learn a generalizable learning algorithm (hence learning to learn) that is optimized to perform on other tasks which are unobserved during meta-training. The latter tasks are referred to as test tasks. After meta-training, the meta-learner is typically allowed to access a few labelled training data points from each test task. In contrast with continual learning, meta-learning is trained in an offline fashion where all the training tasks are available at once to the learner prior to the beginning of meta-training. Moreover, meta-learning does not take catastrophic forgetting into consideration since the performance of a meta-learner is solely assessed based on the test tasks.

## 4. Methods

In this section, we provide an illustration of the seminal Bayesian continual learning algorithms, a corresponding taxonomy, and the principal characteristics of the algorithms in each category of this taxonomy.

### 4.1 Regularization-based Approach

The CL algorithms in this category adopt a strategy based on regularized training for the sake of controlling the parameter updating mechanism throughout the sequentially arriving tasks. In a nutshell, parameters which have a massive impact on prediction are protected against radical changes, whereas the rest of the parameters are allowed to change with more freedom. The premise is that updating parameters via this strategy should potentially achieve a balance between adaptation to new tasks, and mitigating catastrophic forgetting (Li & Hoiem, 2016; Kirkpatrick et al., 2017; Zenke et al., 2017; Zeno et al., 2018; Nguyen et al., 2018; Adel et al., 2020).

The regularization-based approach is innately linked with Bayesian CL. Recall from Section 2.2 that the posterior from the previously encountered task, $t-1$, in Bayesian CL, $p(\boldsymbol{\theta}|\mathcal{D}_{t-1})$, becomes the new prior upon the arrival of (data belonging to) a new task, $t$, (Nguyen et al., 2018; Adel et al., 2020). This paves the way for computing the new posterior $p(\boldsymbol{\theta}|\mathcal{D}_t)$, solely based on the data of the new task $\mathcal{D}_t$, and for designing objectives aiming to potentially alleviate the risk of forgetting the previous knowledge.

Establishing a Bayesian CL model based on exact Bayesian inference is intractable, especially given the fact that the vast majority of continual learners are modelled via deep neural networks. This is the main reason why approximate inference usually becomes a necessity in Bayesian CL frameworks.

#### 4.1.1 MAP ESTIMATION

One of the common strategies in regularization-based Bayesian CL is to base inference within the neural network on regularized maximum likelihood estimation, typically resulting in an objective of the following form:

$$\mathcal{L}^t(\boldsymbol{\theta}) = \sum_{n=1}^{N_t} \log p(\boldsymbol{y}_t^n | \boldsymbol{\theta}, \boldsymbol{x}_t^n) - 0.5\lambda_t(\boldsymbol{\theta} - \boldsymbol{\theta}_{t-1})^T \Sigma_{t-1}^{-1} (\boldsymbol{\theta} - \boldsymbol{\theta}_{t-1}), \tag{9}$$

where $\mathcal{L}^t(\boldsymbol{\theta})$ refers to the online marginal likelihood after having encountered task $t$, $\boldsymbol{\theta}$ depicts the current status of the model parameter values (i.e. after having encountered task $t$), $\boldsymbol{\theta}_{t-1}$ refers to the model parameter values right after having encountered the previous task $t-1$ (and before encountering task $t$), $\lambda_t$ is a hyperparameter which controls the contribution (to the model parameters values) from the previously encountered tasks, and $\Sigma_{t-1}$ is a (usually diagonal) matrix that represents the relative relevance of the regularization term to each element of the model parameters $\boldsymbol{\theta}$. The regularization term, which is the second term at the R.H.S. in (9), pushes the new model parameter estimates $\boldsymbol{\theta}$ towards remaining close to the respective values estimated at the previous task, $\boldsymbol{\theta}_{t-1}$.

The inclusion of the regularization term is considered analogous to maximum a posteriori (MAP) estimation with a Gaussian prior which, at task $t$, is equivalent to $p(\boldsymbol{\theta}|\mathcal{D}_{1:t-1}) = \mathcal{N}(\boldsymbol{\theta}; \boldsymbol{\theta}_{t-1}, \frac{\Sigma_{t-1}}{\lambda_t})$ (Nguyen et al., 2018). In the same vein, a value of $\lambda_t = 0$ is equivalent to performing maximum likelihood estimation since the second term at the R.H.S. in (9) vanishes, whereas the first term is equivalent to the likelihood. In the following, we discuss seminal examples of the MAP regularization approach to Bayesian CL.

**Laplace propagation.** Even though the MAP optimization form outlined in (9) produces a MAP estimate of the model parameters $\boldsymbol{\theta}$, it does not automatically provide an updated estimate of the matrix $\Sigma_t$ for each upcoming task $t$. One solution is to simply assign to it the value of the identity matrix, i.e. $\Sigma_t = I$, but this does not result in an effective treatment of catastrophic forgetting (Goodfellow et al., 2014a; Kirkpatrick et al., 2017). Another (more effective) solution is based on performing Laplace's approximation at each upcoming task (Smola et al., 2004), where the matrix $\Sigma_t^{-1}$ is recursively computed as follows:

$$\Sigma_t^{-1} = \Phi_t + \Sigma_{t-1}^{-1}, \tag{10}$$

$$\text{where } \Phi_t = -\nabla\nabla_{\boldsymbol{\theta}} \sum_{n=1}^{N_t} \log p(\boldsymbol{y}_t^n | \boldsymbol{\theta}, \boldsymbol{x}_t^n)\Big|_{\boldsymbol{\theta}=\boldsymbol{\theta}_t}, \text{ and } \lambda_t = 1$$

At the first task, $t = 1$, the value of $\Sigma_{t-1}^{-1} = \Sigma_0^{-1}$ is initialized using the covariance of the Gaussian prior. Diagonal Laplace propagation, where solely the diagonal terms of $\Sigma_t^{-1}$ are kept, is often adopted since, otherwise, it becomes computationally prohibitive to compute the full Hessian of the likelihood.

**Elastic weight consolidation (EWC).** This is another seminal Bayesian CL algorithm which is based on diagonal Laplace propagation (Kirkpatrick et al., 2017). In the EWC algorithm, regularization is mainly performed via a quadratic penalty that is imposed on the difference between parameter values of the old and new tasks. It also adopts a diagonal Laplace propagation approach, as in (10), yet it differs in the manner pursued to compute the matrix $\Phi_t$. The diagonal of the Fisher information matrix (which is guaranteed to be positive semi-definite) is computed at each task, resulting in a tractable approximation of the average Hessian of the likelihoods. Furthermore, the diagonal of the Fisher information matrix provides an estimate of the importance of the model parameters. The matrix $\Phi_t$ is therefore approximated as follows:

$$\Phi_t \approx \mathrm{diag}\left( \sum_{n=1}^{N_t} \left( \nabla_{\boldsymbol{\theta}} \log p(\boldsymbol{y}_t^n | \boldsymbol{\theta}, \boldsymbol{x}_t^n) \right)^2 \Big|_{\boldsymbol{\theta}=\boldsymbol{\theta}_t} \right) \tag{11}$$

**Synaptic intelligence (SI).** In this algorithm (Zenke et al., 2017), the matrix $\Sigma_t^{-1}$, referred to back in (9), is computed based on quantifying the importance of each parameter per task. Importance measures are computed based on their corresponding contributions to the variation in the global loss. Ultimately, more important parameters are granted less freedom to change. On a high level, this is rather similar to the EWC algorithm in the sense that the values of the parameters which are considered the most influential are further constrained against changes upon encountering new tasks. However, the main difference here is that the parameter importance estimates are computed simultaneously with the task learning procedure, hence there is no need to separately approximate the diagonal of the Fisher information matrix, as is the case with EWC.

**Other related algorithms** Similar algorithms include the work in (Chaudhry et al., 2018) where an online variation of EWC has been proposed, which is also more computationally efficient. The approach therein is based on maintaining a single diagonal Fisher matrix for all tasks, coupled with using a moving average to update the matrix. Furthermore, the moving average also helps make the algorithm a bit less sensitive to the hyperparameter values. Another online version of EWC has been introduced in the algorithm referred to as progress and compress (P&C, Schwarz et al., 2018) which addresses the fact that the EWC Fisher regularizers are exceedingly prone to over-constraining the network parameters. This can ultimately hinder the learning procedure for new tasks in EWC. In addition, they claim that the importance of each previous task in EWC is given an arbitrary scale from the Fisher matrix, which is undesirable. In P&C, this is addressed by normalizing the Fisher information matrix of each task, ultimately leading to an equal treatment of all the previous tasks. A scalable Laplace approximation is presented in (Ritter et al., 2018) based on a block-diagonal, factored approximation (Martens & Grosse, 2015; Botev et al., 2017).

The moments of the posteriors resulting from a Bayesian neural network (BNN) are matched incrementally in the algorithm proposed by Lee et al. (2017). In other words, the moment of the posterior distribution of the BNN which is trained on the first task is matched with the corresponding posterior from the second task, and so on, along with the arrival of each upcoming task. Its basic version works by averaging the parameters of the old BNN and the new BNN (i.e. the BNN parameter values after learning the new task). However, the results of the approximations carried out by the incremental moment matching algorithm are massively conditioned by the search space of the original problem (Foster & Brintrup, 2023); a smooth and convex-like search space is required in order to perform well. The learning rate is adapted according to the estimated level of uncertainty in the probability distribution of the network weights in (Ebrahimi et al., 2020). Inspired by ideas from information theory, the classifier-projection regularization (CPR, Cha et al., 2021) algorithm is based on projecting the conditional probability given by the output of the classifier to a uniform distribution, which is claimed to eventually lead to an improved CL performance. Catastrophic forgetting is addressed via a combination of weight regularization along with gradient projection where gradients of a new task are projected into subspaces that do not interfere with gradients of previous tasks in (Kao et al., 2021).

#### 4.1.2 VARIATIONAL INFERENCE APPROACH

In standard (i.e. non-continual) variational inference, the main objective is to infer an approximate posterior $p(\boldsymbol{\theta}|\mathcal{D})$, which depicts an estimation of the model parameters $\boldsymbol{\theta}$ given a dataset $\mathcal{D}$ where the dataset $\mathcal{D}$ consists of independent and identically distributed (i.i.d.) labelled examples. However, as mentioned above, the i.i.d. assumption does not hold in the CL paradigm. Instead, the learner encounters a set of sequentially arriving tasks each characterized by a dataset $\mathcal{D}_t = \{\boldsymbol{x}_t^n, \boldsymbol{y}_t^n\}_{n=1}^{N_t}$, where $t \in \{1, 2, \ldots, m\}$ is the task index, and $N_t$ is the size of the training dataset of task $t$. Even though the data belonging to each task follows the i.i.d. assumption (i.e. solely among data points belonging to the same task) , this is not generally the case for data belonging to different tasks.

**Variational continual learning (VCL).** The variational inference approach in CL is meticulously represented by variational continual learning (VCL, Nguyen et al., 2018; Loo et al., 2021) where the approximate posterior at the previous task $t-1$ is then treated as the prior at the current task $t$, and a Kullback-Leibler divergence (KL-divergence) term is used to regularize the objective. More specifically, after encountering $t$ tasks, the exactly intractable posterior $p(\boldsymbol{\theta}|\mathcal{D}_{1:t})$ is approximated via a tractable variational distribution $q_t(\boldsymbol{\theta})$ which can be described as follows:

$$q_t(\boldsymbol{\theta}) = \underset{q \in Q}{\operatorname{argmin}} \, \mathrm{KL}\Big(q(\boldsymbol{\theta}) \parallel \frac{1}{Z_t} q_{t-1}(\boldsymbol{\theta}) \, p(\mathcal{D}_t|\boldsymbol{\theta})\Big), \quad \text{for } t = 1, 2, \ldots, m \tag{12}$$

where $Q$ is a set of approximate posteriors, and $p(\mathcal{D}_t|\boldsymbol{\theta}) = \prod_{n=1}^{N_t} p(\mathbf{y}_t^n|\boldsymbol{\theta}, \mathbf{x}_t^n)$ is the likelihood which depicts the information we can obtain from the data of the current task. The very first prior, which is defined as the approximate distribution at time step 0, i.e. prior to encountering any tasks, is $q_0(\boldsymbol{\theta}) = p(\boldsymbol{\theta})$. Note that the normalizing constant, $Z_t$ for task $t$, which is equivalent to computing $q_{t-1}(\boldsymbol{\theta})p(\mathcal{D}_t|\boldsymbol{\theta})$ over all the possible values of the model parameters $\boldsymbol{\theta}$, is intractable. Nonetheless, it is not required for inferring the optimal approximate posterior $q_t(\boldsymbol{\theta})$ since it does not depend on the model parameters $\boldsymbol{\theta}$ (which are marginalized out therein), but only on the input data $\mathcal{D}$ which do not represent any degrees of freedom in the optimization process. The VCL framework allows the approximate posterior of the current task $q_t(\boldsymbol{\theta})$ to be updated incrementally given the approximate posterior of the previous task $q_{t-1}(\boldsymbol{\theta})$ in an online fashion.

Approximating the posterior distribution according to the approach defined in (12) is equivalent to maximizing the variational evidence lower bound (ELBO) on the online marginal likelihood $\mathcal{L}_{\mathrm{VCL}}^t(q_t(\boldsymbol{\theta}))$ for the current task $t$ which is denoted as follows:

$$\mathcal{L}_{\mathrm{VCL}}^t(q_t(\boldsymbol{\theta})) = \sum_{n=1}^{N_t} \mathbb{E}_{\boldsymbol{\theta} \sim q_t(\boldsymbol{\theta})} \left[\log p(\boldsymbol{y}_t^n|\boldsymbol{\theta}, \boldsymbol{x}_t^n)\right] - \mathrm{KL}(q_t(\boldsymbol{\theta}) || q_{t-1}(\boldsymbol{\theta})). \tag{13}$$

The first term at the R.H.S. in (13) is the expected log-likelihood of the CL model over the dataset of the current task $t$, whereas the second term aims to penalize the difference between the current approximate posterior and its counterpart at the previous task. The whole lower bound expression in (13) cannot be computed in closed form in CL. As a result, gradient computations of this variational objective usually involve the employment of a local reparameterization trick along with a simple Monte Carlo (Salimans & Knowles, 2013; Kingma & Welling, 2014; Kingma et al., 2015; Nguyen et al., 2018). Unlike the methods adopting the MAP estimation approach, such as EWC and SI, VCL involves fewer free parameters that need tuning on a validation set. This is effectively advantageous since an excessive number of free parameters, along with the respective validation set, can be onerous in the online setting (Nguyen et al., 2018).

**Other related algorithms** Other variational inference-based algorithms include the work in (Joseph & Balasubramanian, 2020) where a task-specific meta-distribution of the network weights is learned by training an ensemble of models per task, and then using such an ensemble as a training set for a task-conditioned variational autoencoder (VAE, Kingma & Welling, 2014; Kingma et al., 2014). For each task, the VAE is conditioned on the respective task-specific prior. VAEs have as well been utilized in a CL context in (Egorov et al., 2021) where an optimal prior is first defined for the CL VAE, and then an optimal additive expansion of the current prior is inferred to match every upcoming task. The method in (Ahn et al., 2019) addresses the excessive memory requirements of both the Fisher information matrix-based approach adopted by EWC and the variational inference-based approach adopted by VCL (where there is at least one variance term associated with each weight parameter) via associating the learnable variances with the hidden units of the neural network, rather than with the network weight parameters. It is argued in (Chen et al., 2019) that natural gradient methods (Pascanu & Bengio, 2014) can be a better choice than conventional gradient descent since the former gives the direction of the steepest descent in the Riemannian (rather than Euclidean) space, which means that natural gradients would favor smaller changes, in terms of the parameter distributions, while performing the optimization. Smaller changes are preferred since this ultimately leads to a CL model which is less prone to catastrophic forgetting.

### 4.2 Replay-based Approach

This approach relies on storing or replaying data from previous tasks, which leads to overheads for the model such as data storage, replay, and/or optimization to select (or generate) the data points (Titsias et al., 2020). Storing data from previous tasks can be tricky in practice due to security and/or privacy concerns. Other methods are based on learning a generative model to generate data points from previous tasks (Farquhar & Gal, 2019). This approach results in less overhead in terms of storage. However, there is the added overhead of training the generative model. We next shed light on the most notable Bayesian CL algorithms of the replay-based approach.

**Variational generative replay (VGR).** The VGR algorithm (Farquhar & Gal, 2019) represents the replay-based counterpart of VCL. VGR has been developed as a likelihood-focused Bayesian alternative to the prior-focused VCL. Rather than relying on the posteriors belonging to earlier tasks, the VGR CL model is adapted via continuously changing the likelihood term. According to the terminology developed in (Farquhar & Gal, 2019), VCL is considered prior-focused because it continually treats the old posterior as the prior upon the arrival of a new task. The VGR approach is much more expensive than VCL mainly since a generative adversarial network (GAN, Goodfellow et al., 2014b; Goodfellow, 2016) is trained at every task using the respective training data. Moreover, the GANs of all the previous tasks are stored and then sampled to create the replayed data which is to be used for the upcoming task.

The VGR algorithm basically rejects the assumption made by VCL in (13) which notes that the posterior from the previous task $t-1$, $q_{t-1}(\boldsymbol{\theta})$, that is used as the prior for the upcoming task $t$, is a good initial point for learning to approximate the CL posterior $q_t(\boldsymbol{\theta})$. Instead, in VGR, successive datasets are trained via GANs which are stored and eventually used to simulate datasets belonging to the old tasks.

In order to capture this behavior in a Bayesian setting, the standard ELBO is expanded to account for the multiple previous datasets (i.e. those belonging to the old tasks):

$$\mathcal{L}^t_{\mathrm{VGR}}(q_t(\boldsymbol{\theta})) = \sum_{i=1}^{t}\left[\sum_{n=1}^{N_i} \mathbb{E}_{\boldsymbol{\theta}\sim q_i(\boldsymbol{\theta})}\left[\log p(\boldsymbol{y}_i^n|\boldsymbol{\theta},\boldsymbol{x}_i^n)\right]\right] - \mathrm{KL}(q_t(\boldsymbol{\theta})||p(\boldsymbol{\theta})). \tag{14}$$

The first term at the R.H.S. in (14), which depicts the expected log-likelihood of VGR, is summed over all the previously encountered tasks $i = 1, 2, \ldots, t-1$, as well as the upcoming task $t$. Furthermore, the KL-divergence is computed between the estimate of the current posterior $q_t(\boldsymbol{\theta})$ and the original prior $p(\boldsymbol{\theta})$.

Given the fact that the previous datasets, $\mathcal{D}_i = \{\boldsymbol{x}_i^n, \boldsymbol{y}_i^n\}_{n=1}^{N_i}$, for $i = 1, 2, \ldots, t-1$, are no longer available in CL, the log-likelihood in (14) cannot be computed directly. As such, the expectation of the log-likelihood for each previous task $i, i < t$, is estimated separately as follows:

$$\sum_{n=1}^{N_i} \mathbb{E}_{\boldsymbol{\theta}\sim q_i(\boldsymbol{\theta})}\left[\log p(\boldsymbol{y}_i^n|\boldsymbol{\theta},\boldsymbol{x}_i^n)\right] \approx \int \log\big[p(\boldsymbol{y}|\boldsymbol{\theta},\boldsymbol{x})\big]p_i(\boldsymbol{x},\boldsymbol{y})q(\boldsymbol{\theta})\, d\boldsymbol{x}d\boldsymbol{y}d\boldsymbol{\theta}. \tag{15}$$

This form of the log-likelihood relies on a separately trained GAN $p_i(\boldsymbol{x},\boldsymbol{y})$. Prior to the beginning of the training procedure for every upcoming task $t$, the stored GANs of the previous tasks $i, i < t$, are utilized to produce samples $\hat{\boldsymbol{x}}, \hat{\boldsymbol{y}} \sim p_i(\boldsymbol{x},\boldsymbol{y})$. Thus, the upcoming task $t$ is ultimately trained on the composite dataset consisting of the data which is generated from all the previous tasks as well as the current dataset of task $t$. In other words, the upcoming task $t$ is trained on $\tilde{\mathcal{D}_t} \equiv (\tilde{\boldsymbol{x}}, \tilde{\boldsymbol{y}}) = (\hat{\boldsymbol{x}} \cup \boldsymbol{x}_t, \hat{\boldsymbol{y}} \cup \boldsymbol{y}_t)$. Samples drawn from the dataset $\tilde{\mathcal{D}_t}$ are then used to approximate the expected log-likelihood, which is in turn used to compute the VGR ELBO loss in (14).

**Coreset VCL.** The coreset version of VCL (Nguyen et al., 2018) aims to address the fact that the aforementioned standard VCL is suffering from performing repeated approximations over the subsequent tasks which potentially makes the algorithm increasingly prone to the risk of catastrophically forgetting the old tasks. The coreset version of VCL mitigates this issue via storing a subset of the data belonging to each of the previous tasks, and then replaying such a subset when learning future tasks. The subset of the old data from each previous task (referred to as the coreset) is thus utilized to refresh the memory of the VCL model about the old tasks, hence potentially mitigating the risk of catastrophic forgetting. Apparently, this comes at the cost of the computational and storage constraints associated with continually selecting, storing and then replaying the coresets.

For each task, the new coreset $C_t$ is produced by selecting a subset of the data points belonging to the upcoming task $t$, and adding this subset to the previous selections already belonging to the old coreset $C_{t-1}$. Heuristics are usually used to select the data points. One heuristic is to randomly select $K$ data points from the respective dataset $\mathcal{D}_t$, and then add such points to the old coreset $C_{t-1}$ to form the current coreset $C_t$. Another alternative is to employ the greedy $K$-center algorithm (Gonzalez, 1985) to return $K$ data points per task. The latter option incurs further computational cost, albeit with the advantage of obtaining coreset points, per task, which are further spread throughout the input space.

Upon encountering a new task $t$, and the corresponding dataset $\mathcal{D}_t$, a subset of such data is added to the coreset, resulting in $C_t$. Let's refer to the rest of the data points in $\mathcal{D}_t$ as the non-coreset data points. The non-coreset points are first used to update the online variational distribution as follows:

$$\tilde{q}_t(\boldsymbol{\theta}) \leftarrow \underset{q \in Q}{\operatorname{argmin}} \, \mathrm{KL}\Big(q(\boldsymbol{\theta}) \parallel \frac{1}{\tilde{Z}_t} \tilde{q}_{t-1}(\boldsymbol{\theta}) \, p(\mathcal{D}_t \cup C_{t-1} \setminus C_t | \boldsymbol{\theta})\Big), \tag{16}$$

where $\tilde{q}_t(\boldsymbol{\theta})$, for task $t$, refers to the variational distribution which is continually updated, once per task. The data $\mathcal{D}_t \cup C_{t-1} \setminus C_t$ refers to the combination of: i) the old coreset $C_{t-1}$, and ii) the data points of the new dataset $\mathcal{D}_t$ which have not been selected as part of the coreset points for task $t$, $\mathcal{D}_t \setminus C_t$. A recursive variational process is therefore applied to decompose the contributions of the current dataset $D_t$ into the points which have not been selected into the coreset, $\mathcal{D}_t \setminus C_t$, and the selected coreset points from task $t$. This results in redundancy-free computations, i.e. no data point is utilized twice in inferring the variational posterior. The coreset points from task $t$ are then used to infer the final variational distribution for task $t$, which is only used for prediction, and not for iterative propagation:

$$q_t(\boldsymbol{\theta}) \leftarrow \underset{q \in Q}{\operatorname{argmin}} \, \mathrm{KL}\Big(q(\boldsymbol{\theta}) \parallel \frac{1}{Z_t} \tilde{q}_t(\boldsymbol{\theta}) \, p(C_t | \boldsymbol{\theta})\Big). \tag{17}$$

Similar to the vanilla VCL case outlined in (12), the normalizing constants $\tilde{Z}_t$ and $Z_t$ are not required to infer the optimal approximate posteriors $\tilde{q}_t(\boldsymbol{\theta})$ and $q_t(\boldsymbol{\theta})$, respectively.

**Functional regularization for continual learning using Gaussian processes (FRCL).** The FRCL algorithm (Titsias et al., 2020) bases its Bayesian inference over the function space rather than the parameters of a deep neural network. An approximate posterior belief is memorized over the underlying function space which relies on a Gaussian process obtained by treating the weights of the last layer of the neural network as Gaussian distributed.

In FRCL, the classification model is a deep neural network which is denoted as $f_t(\boldsymbol{x}; \boldsymbol{\theta}) = w_t^T \phi(\boldsymbol{x}; \boldsymbol{\theta})$, where $\boldsymbol{\theta}$ are the model parameters, and $t$ is the task index. According to their model, the parameters $\phi(\boldsymbol{x}; \boldsymbol{\theta})$ depict the representation which is shared across all tasks, whereas $w_t$ is the vector of task-specific weights (for task $t$). A distribution is obtained over functions by placing a Gaussian prior over the weights, $w_t \sim \mathcal{N}(0, \sigma_w^2 I)$, where $\sigma_w$ is the standard deviation. There is ultimately a specific weight vector $w_t$ for each task $t$, but the aforementioned distribution refers to an infinite set of tasks that can be handled by the same shared representation $\phi(\boldsymbol{x}_t; \boldsymbol{\theta})$. In other words, each task-specific weight vector can be considered an independent draw from the GP (Rasmussen & Williams, 2006):

$$f_t(\boldsymbol{x}) \sim \mathcal{GP}(0, k(\boldsymbol{x}, \boldsymbol{x}')), \tag{18}$$

$$k(\boldsymbol{x}, \boldsymbol{x}') = \sigma_w^2 \phi(\boldsymbol{x}; \boldsymbol{\theta})^T \phi(\boldsymbol{x}'; \boldsymbol{\theta}), \tag{19}$$

where the kernel function is defined by the dot product of the shared representation $\phi$.

The memorization is achieved using inducing points that are formed via a sparse Gaussian process (GP). Inducing points denote a fixed-size subset of the data points of each task, which is selected such that it optimally represents the task. These subsets are then used to regularize the learning process of future tasks through KL-divergence terms in a form similar to the VCL formulation illustrated in (13), but based on the inducing points. The FRCL algorithm can be seen as a replay-based as well as regularization-based method. It therefore suffers from the optimization overhead needed to obtain the inducing points based on sparse GPs. Furthermore, there is also the scalability issue stemming from the fact that the inducing points should not only be optimized for, but also stored, for each task.

The FRCL algorithm (Titsias et al., 2020) has as well provided the basis for subsequent related methods, e.g. (Pan et al., 2020; Rudner et al., 2022). In (Pan et al., 2020), they use a Laplace approximation and form a Gaussian process by forcing the corresponding kernel to employ all the weights of the neural network (rather than solely the last layer). Rather than using a Laplace approximation, the method in (Rudner et al., 2022) directly performs an optimization of the variance parameters via variational inference.

**Other related algorithms** Other GP-based CL algorithms include variational auto-regressive Gaussian processes (VAR-GPs, Kapoor et al., 2021) where an auto-regressive variational distribution is used along with inducing points to memorize the old tasks. A complementary memory of raw data belonging to old tasks is utilized along with a variational posterior approximating the weights of a BNN by a diagonal Gaussian distribution in (Kurle et al., 2020). Online memory selection based on an information-theoretic criterion is introduced in (Sun et al., 2022).

Other replay-based examples include the work in (Borsos et al., 2020) where coresets that summarize old tasks are constructed via cardinality-constrained bi-level optimization which they solve by greedy forward selection based on weighted data summarization. The work in (Lyu et al., 2023) applies a probabilistic strategy to adjust inter-task weights so as to adapt batch normalization, which they use along with a memory buffer. They assume that the statistics of each batch for a certain layer follow a probability distribution where the randomness originates from the values of the neural network parameters. A VAE-based memory buffering approach is introduced in (Ye & Bors, 2022a) which consists of a short-term memory (STM) that continually stores recent samples, along with a long-term memory (LTM) that aims to preserve a wide diversity within all of the stored samples. Certain samples can be transferred from the STM to the LTM according to an information diversity selection criterion. A single shared meta-distribution is learned in (Henning et al., 2021) where the parameters of all the previous task-specific posteriors are replayed, but it suffers from the limitation of incurring a significant computational overhead.

### 4.3 Architecture-based Approach

In this approach, the main objective is to aim at controlling the tradeoff between stability and adaptation by dividing the architecture into two types of components: i) global parts which are reusable and less prone to changes throughout the tasks, and ii) task-specific parts which are specifically devoted to individual tasks. One of the advantages of this incremental structure learning approach to CL is the ability to mitigate catastrophic forgetting via allowing room for task-specific parameters that can improve the performance of the model on the respective tasks, without affecting the rest of the tasks since the latter are protected by the restrained control of the shared parameters. On the other hand, scalability issues could be a hindrance to adopting the architecture-based approach with a large number of CL tasks due to the potentially prohibitive expansion in the size of the model's architecture.

**Continual learning with adaptive weights (CLAW).** Another algorithm which is based on VCL, yet with automatic architecture adaptation capabilities, is the algorithm referred to as continual learning with adaptive weights (CLAW, Adel et al., 2020). Rather than adopting the commonly rigid multi-task architecture with a fixed division between the shared components at the bottom and task-specific components at the top, CLAW adaptively identifies which parts of the network to share via a flexible data-driven approach based on variational inference, which ultimately results in optimizing the tradeoff between mitigating catastrophic forgetting and improving task transfer.

In CL scenarios involving heterogeneous tasks, the data-driven approach adopted by CLAW is capable of determining the appropriate amount of sharing across tasks. This is achieved via automating the architecture adaptation process by either keeping each hidden neuron intact (i.e. acting as a global neuron), or locally adapting the neuron to each new task. The latter course (i.e. local adaptation) is flexibly achieved via learning the range within which the adaptation parameters can vary. For each neuron, a total of three parameters are added. One binary parameter is responsible for learning whether or not the respective neuron should be adapted. In addition, two other parameters control the magnitude of adaptation such that the degree of adaptation is also learned. All parameters are learned via variational inference. Another advantage of CLAW is the fact that there is no significant expansion of the architecture itself throughout the whole process, since no new neurons are added. Moreover, there is no need to store (nor generate) data from previous tasks.

**Indian Buffet neural networks.** An Indian Buffet process (IBP) prior is placed over the structure of a BNN in (Kessler et al., 2021), which ultimately results in a model that is capable of dynamically adapting the complexity the BNN according to the amount of encountered data. This IBP-based CL model utilizes online variational inference with reparameterizations of Bernoulli and Beta distributions to form the IBP prior.

The IBP prior (Griffiths & Ghahramani, 2011) is assigned as a prior on an infinite, binary latent matrix $Z$. This permits the model to perform inference on how many hidden units (i.e. neurons), per layer, are required for each task. The weights of the BNN are then treated as random draws from non-interacting Gaussians (Blundell et al., 2015). This is followed by iteratively applying variational Bayesian updates in a VCL-like fashion.

Refer to the data matrix of task $t$ (excluding the class labels) as $\boldsymbol{X}_t \in \mathbb{R}^{N_t \times d_t}$ where $N_t$ is the size of the training dataset of task $t$, and $d_t$ is the number of features for each data point of such a dataset. Based on matrix decomposition, the data matrix $\boldsymbol{X}_t$ can be represented as a combination of latent features in the following manner: $\boldsymbol{X}_t = ZA + \epsilon$, where $Z \in \mathbb{Z}_2^{N_t \times k}$, $A \in \mathbb{R}^{k \times d_t}$, $\epsilon$ is an observation noise, and $\mathbb{Z}_2 = \{0, 1\}$. Each element in the matrix $Z$ represents information about the presence or absence of a latent feature from the matrix $A$. For instance, $z_{ik} = 1$ refers to the presence of the corresponding latent feature $A_k$ in data point $\boldsymbol{X}_i$.

An IBP prior is then placed over $Z$ in (Kessler et al., 2021; Kumar et al., 2021) based on the stick-breaking formulation (Teh et al., 2007). Each column $z_k$ is assigned a probability $\pi_k$. The probability of selecting a latent feature is then identified according to a Bernoulli distribution: $z_{ik} \sim \text{Bern}(\pi_k)$. The parameter $\pi_k$ in turn results from the following stick-breaking process: $\pi_k = \prod_{i=1}^{k} v_i$, where $v_i \sim \text{Beta}(\alpha, 1)$. It can clearly be seen that the value of $\pi_k$ decreases exponentially with $k$. The concentration parameter $\alpha$ of the Beta distribution controls the overall number of expected latent features; a large $\alpha$ ultimately leads to the presence of more latent features. The binary matrix $Z$ is then applied element-wise to each hidden layer, ultimately resulting in controlling the number of neurons per layer according to a variational objective involving the aforementioned stick-breaking process.

Few other Bayesian CL algorithms are architecture-based. One more example can be found in (Wang et al., 2021) which dynamically expands parameters when learning each new task, and then selectively combines the expanded parameters. It is argued that this method is consistent with the underlying mechanism of biological active forgetting. Another example of a CL algorithm where an IBP prior is utilized in a Bayesian non-parametric approach based on the stick-breaking formulation can be found in (Mehta et al., 2021). The IBP prior therein is paired with a factorization of the weight matrices of the neural network in a manner that allows for factor reuse, and consequently positive knowledge transfer, among the different tasks.

### 4.4 Class-Incremental Learning

Unlike the case with the task-incremental learning setting, very few CIL algorithms base their online updating on ideas stemming from Bayesian inference and/or learning. It is important to stress on the fact that what we refer to as Bayesian CL in this manuscript mainly depicts the adoption of a Bayesian approach in the *online updating* process required for CL frameworks, e.g. with VCL where the old posterior becomes the new prior, etc. In other words, we do *not* refer to frameworks that solely apply a Bayesian classifier to learn from each task (in isolation), without performing Bayesian inference on the underlying CL model parameters $\boldsymbol{\theta}$.

**Few-shot class-incremental learning adaptation via latent variable models (CIAM)** As such, in the strict sense of this definition, the FSCIL algorithm introduced in (Adel, 2025) can so far be considered the only Bayesian CIL algorithm in the literature. Compared to the standard CIL setting, an added level of difficulty arises when the training data belonging to most classes are scarce, as in FSCIL where the classes belonging to the first task have more data than the classes belonging to every subsequent task (i.e. after the first). In addition to the risk of catastrophic forgetting, such data scarcity compounds other issues in the FSCIL setting, like bias and overfitting. The CIAM algorithm adapts the representations of the few-shot classes (i.e. classes belonging to every task after the first task -from the second task onwards), and balances them with relevant previous knowledge based on a latent variable model that is designed as a form of a VAE that is bespoke to mitigate the bias and catastrophic forgetting that can occur in FSCIL. When learning the few-shot classes, the amortization characteristics of the VAE introduced in (Adel, 2025) are leveraged to adapt not only the current few-shot class, but also the relevant previously encountered classes.

In the following we shed light on two algorithms which are loosely close to being Bayesian CIL algorithms (although, again, not strictly speaking):

**Gaussian process tree.** Inference in Gaussian processes (GPs) can be computationally challenging, especially with large datasets. Such challenges are further compounded in the CIL setting. The work in (Achituve et al., 2021) tackles these challenges by introducing a method for multi-class classification based on a tree-based hierarchical model of GPs in which each internal node of the tree fits a GP to the data using the Pòlya-Gamma augmentation scheme (Polson et al., 2013). According to the taxonomy adopted above for task-incremental learning, this algorithm can be considered a replay-based algorithm due to the utilization of inducing points that act as a proxy for the training data of the previous tasks.

Prior to this work, the aforementioned Pòlya-Gamma augmentation scheme enabled inference for binary classification via conditioning the GP posterior on an augmented Pòlya-Gamma variable. The work in (Achituve et al., 2021) extends this treatment to multi-class CIL classification. They base their solution on a tree-based model in which each node maps to a binary classification task using a GP along with inducing points from every previous task. Even though the tree of binary-classification GPs warrants a CIL framework, the number of binary classifiers involved in the resulting hierarchical classifier, along with the optimization required for each GP (i.e. each node) as well as the optimization associated with the overall tree, can become prohibitively massive with a large number of classes, which ultimately leads to computational complications.

**CIL with generative classifiers.** The method developed in (van de Ven et al., 2021) learns from the data of each class based on a generative (rather than a discriminative) classifier. In other words, rather than learning from the conditional distribution $p(\boldsymbol{y}|\boldsymbol{x})$, they learn the joint distribution $p(\boldsymbol{x}, \boldsymbol{y})$ for each class, factorize it as $p(\boldsymbol{x}|\boldsymbol{y})p(\boldsymbol{y})$, and then perform classification based on Bayes' rule. One of the several scalability issues associated with this method is the fact that they train a separate VAE for each and every encountered class. Furthermore, an expensive importance sampling procedure is needed to estimate the respective likelihood $p(\boldsymbol{x}|\boldsymbol{y})$. As noted in (van de Ven et al., 2021), for each class, no less than $10,000$ importance samples are needed in order to achieve an acceptable degree of fidelity with estimating the likelihood per class.

## 5. Connection with Developmental Psychology

Humans are more adept than machines at learning continually. Unlike machines, (healthy) humans hardly suffer from issues like forgetting important information acquired in the past solely due to the acquisition of more recent information. In other words, issues such as catastrophic forgetting and the stability-plasticity tradeoff are instinctively addressed in an efficient manner by the human brain (Finn et al., 2017; Rostami et al., 2020).

There have previously been attempts to investigate the unequivocal supremacy of humans at learning continually (Hadsell et al., 2020; McCaffary, 2021). Yet, the bulk of such explorations solely focussed on the perspective of neuroscience. We offer a fresh perspective herein which is based on developmental psychology. We aim to gain further insights for CL via analyzing concepts closely associated with memory, forgetting, and the remarkable ability of humans to continually adapt to different situations throughout their lives.

Developmental psychology focuses on studying how humans adapt to different types of changes (such as cognitive, social, emotional and intellectual) throughout their life span (Grotuss et al., 2007; Greenfield, 2009). Even though the CL challenges encountering humans are more tricky than those encountering machines, given the fact that changes taking place during the life span of an individual comprise various scopes and natures (e.g. cognitive, personal, etc), the human capability to adapt to such onerous changes remains superior. This is one of the primary reasons why the developmental psychology perspective can be inspiring for CL.

### 5.1 Similarities between how Humans and Machines Learn Continually

One of the few similarities that exist between human CL and automated CL is the fact that the ability of both humans and machines to learn continually turn to decline as long as the number of tasks they encounter keep increasing. Notwithstanding their supremacy to learn continually, humans become less able to learn, remember, and adapt to dynamic environments when getting older. Moreover, the levels of plasticity of humans show a consistent tendency to decrease along with increasing age (Hensch, 2004). The increasing age necessarily implies that the respective humans encounter more tasks where they are required to both gain new knowledge as well as keep the knowledge acquired from tasks encountered throughout their previous years. This is similar to what happens with machines which become further prone to catastrophic forgetting as well as decreasing levels of plasticity along with an increasing number of tasks.

Another aspect of similarity is what is referred to as scaffolding (Margolis, 2020) in child development. When a child encounters a new task, which in this case is referred to as the zone of proximal development (Margolis, 2020), adults play a role in helping children to acquire new skills that build on the knowledge they already have in the past, such that children can ultimately accomplish the new task successfully. We therefore conjecture that scaffolding can be considered a human way of enforcing positive transfer (in CL terms) and of facilitating adaptation to new information without negatively affecting the already acquired knowledge.

### 5.2 Links between the Stability-Plasticity Tradeoff and Developmental Psychology

In developmental psychology, cognitive flexibility refers to the ability of a human being to adapt their thinking and behaviour to changing situations or demands, which often involves switching among different tasks or concepts (Canas et al., 2006). On the other hand, cognitive stability refers to the ability to maintain focus on a task despite distractions, particularly when faced with competing tasks and/or environmental changes (Canas et al., 2006).

An assumption that holds for several (yet not all) schools of thought in developmental psychology is that increased cognitive flexibility may come at the expense of cognitive stability (Canas et al., 2003). Conversely, increased cognitive stability may reduce cognitive flexibility. As such, according to such a school of thought, the stability-flexibility tradeoff in developmental psychology can be conceived as a continuous spectrum with two extremes. One extreme is depicted by individuals who are extremely flexible, yet unable to maintain focus on one task for a considerable amount of time. The other extreme is represented by individuals who maintain focus on one task at a time, yet might struggle to adapt to unexpected changes prior to finalizing the task in hand. Hence, this school of thought conceptualizes the stability-flexibility tradeoff in developmental psychology in a manner that is rather equivalent to its counterpart in CL (the stability-plasticity tradeoff) with two competing goals between which there is a potential tension, and where achieving one potentially necessitates a reduction in the other. In other words, it is impossible to achieve optimality with both goals since enhancing one aspect (stability or flexibility) would negatively impact the other. In developmental psychology, one of the principal proponents of this idea is the control dilemma theory (Qiao et al., 2023) which suggests a reciprocal relationship between cognitive stability and cognitive flexibility, and notes that achieving one goal often requires sacrificing the other.

Having said that, one major difference between the realm of developmental psychology and that of CL is that there is another school of thought within the former, which indicates that cognitive stability and cognitive flexibility can indeed be regulated independently (Egner, 2023). According to this doctrine, individuals do not necessarily have to balance the need to maintain focus on one particular task with the need to switch to a new task. This is due to several factors among which the fact that there are individual differences among humans (unlike machines) where some people may be instinctively capable of maintaining focus while switching from one task to another with no performance reduction in either. Such people would therefore thrive in environments that require frequent switching and adaptation. In such a case, the tradeoff between cognitive stability and cognitive flexibility is not obligatory to optimize performance. In other words, unlike machines that aim to learn continually, it is possible for an individual to be both stable and flexible at the same time (Geddert & Egner, 2022).

Another factor that negates the necessity of the tradeoff between cognitive stability and cognitive flexibility is the capacity of the human brain to intelligently adapt to changing demands, and to be flexible in some situations, while being stable in others. This means that humans possess what is referred to as contextual adaptation (Siqi-Liu & Egner, 2020) that enables them to prioritize one aspect (out of cognitive stability or flexibility) over the other according to the situation. This can be explained in simpler terms via referring to the human innate ability to strategically prioritize either stability or flexibility for the sake of obtaining the best possible outcome at any given situation, without having to create tension between the two concepts (i.e. between stability and flexibility). For instance, a surgeon who must maintain a steady hand and focus while performing a delicate operation would then be adept at prioritizing stability until finalizing the operation. Another example is when an individual accepts to voluntarily forget information since they know for certain it will no longer be relevant for their development. For example, if a particular piece of knowledge that was acquired in the past contradicts with new knowledge where the latter is believed to be correct, the brain would be more likely to forget the old (and wrong) knowledge. One other factor that invalidates the aforementioned tradeoff with humans is that (unlike with machines) forgetting can have a positive aspect with humans since it plays a fundamental role in refining the knowledge acquisition process (Martínez-Plumed et al., 2015).

## 5.3 Forgetting

In developmental psychology there is a total of a five popular theories of forgetting (Brainerd et al., 1990; Mcleod, 2023). We focus here on the ones from which we can draw analogies with CL.

The most striking analogy with CL can be illustrated according to one of the five theories, which is referred to as the *interference theory of forgetting*. According to this theory, retrieval of previous knowledge can be interrupted by new information. For example, on Friday, an individual may not recall what their breakfast contained on Sunday due to the number of meals consumed in between (Brainerd et al., 1990). In this regard, two contrasting concepts, referred to as retroactive interference and proactive interference, are closely linked with this interpretation of forgetting. As illustrated in Figure 5, retroactive interference takes place when new memories disrupt old memories, like the breakfast example above. In contrast, proactive interference refers to the phenomenon where old memories interfere with new memories (i.e. the other way round) (Ebert & Anderson, 2009). We illustrate both phenomena with an example in Figure 6 with an individual who had learned Italian in the past and is currently learning French. In such a case, retroactive interference refers to the case when the individual encounters difficulties recalling Italian words due to the more recent experience of learning French. On the other hand, when the individual aims to speak French, yet finds out that their previous knowledge of Italian interferes with French, this is an occurrence of the proactive interference phenomenon. Both phenomena of proactive and retroactive interference are thought to be more likely to take place with similar memories (similar to the aforementioned example with Latin languages). For example, old and new telephone numbers are more likely to interfere with one another than other numbers stored in the memory (e.g. apartment numbers, etc). Similarly, students who study similar subjects at the same time experience interference more often than when studying diverse subjects (Chandler, 1991).

**Retroactive interference**

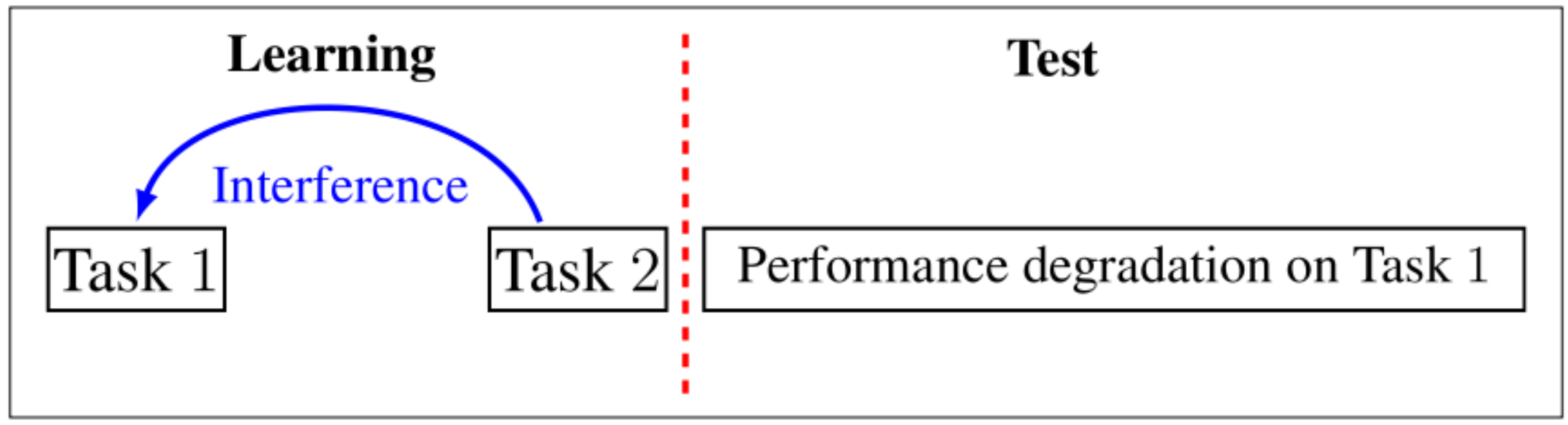

**Proactive interference**

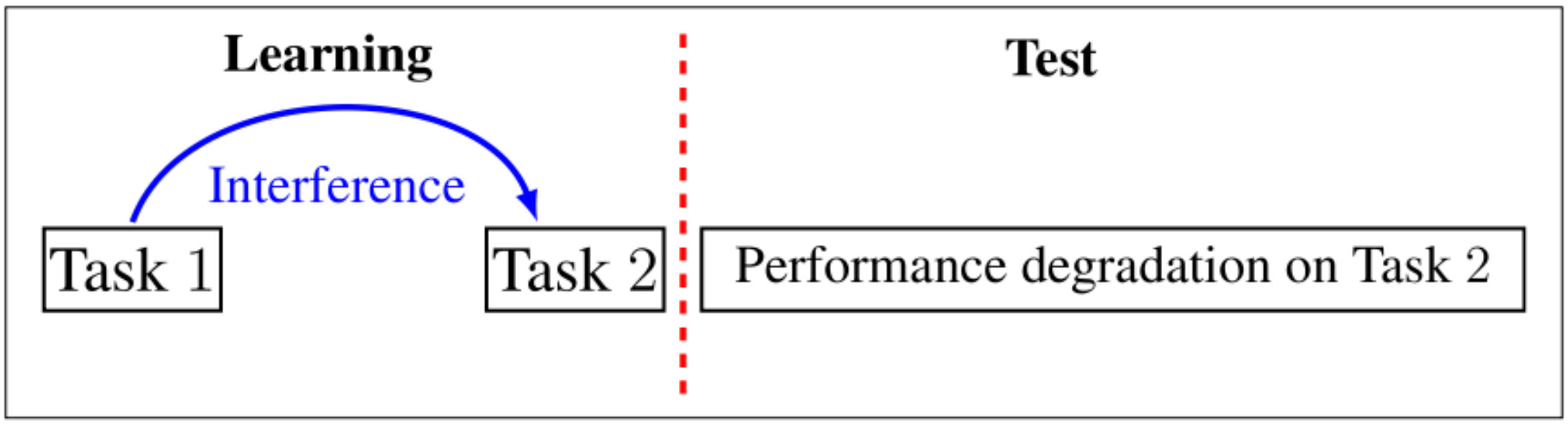

Figure 5: Retroactive interference and proactive interference in developmental psychology. Retroactive interference occurs when new memories interfere with old memories. In contrast, proactive interference refers to the phenomenon where old memories disrupt new memories (i.e. the other way round). An explanation with an example is to follow in Figure 6.

The current research direction in CL is quite analogous with the *displacement theory of forgetting*. Current trends in CL are circulated around balancing forgetting with plasticity via the stability-plasticity tradeoff. Similarly, the displacement theory of forgetting (Mcleod, 2023) strongly correlates forgetting with the short-term memory which has limited retaining capacity, and can solely maintain a limited amount of information. Therefore, the displacement theory of forgetting studies issues like which information is retained in the fixed-size short-term memory, and how to focus attention to retain the important knowledge amid these limited-capacity conditions.

In contrast, the *retrieval theory of forgetting* interprets forgetting differently. According to the retrieval theory, forgetting of information does not necessarily mean that such information is completely lost. Forgetting can as well occur due to the individual's inability to retrieve information from their (long-term) memory (Spear, 1971). Information can be stored in the long-term memory, i.e. it is not completely lost, yet the individual does not manage to retrieve it at a given moment. A common example is where an individual fails to recall a particular word during a specific conversation, even though they have the feeling that this word is stuck at the tip of the tongue. It is often the case that the word can actually be recalled a few hours later. As such, this knowledge that was not recalled when needed earlier is not completely lost; this was solely a retrieval failure. This retrieval interpretation of forgetting can hardly be associated with the current directions in CL research.

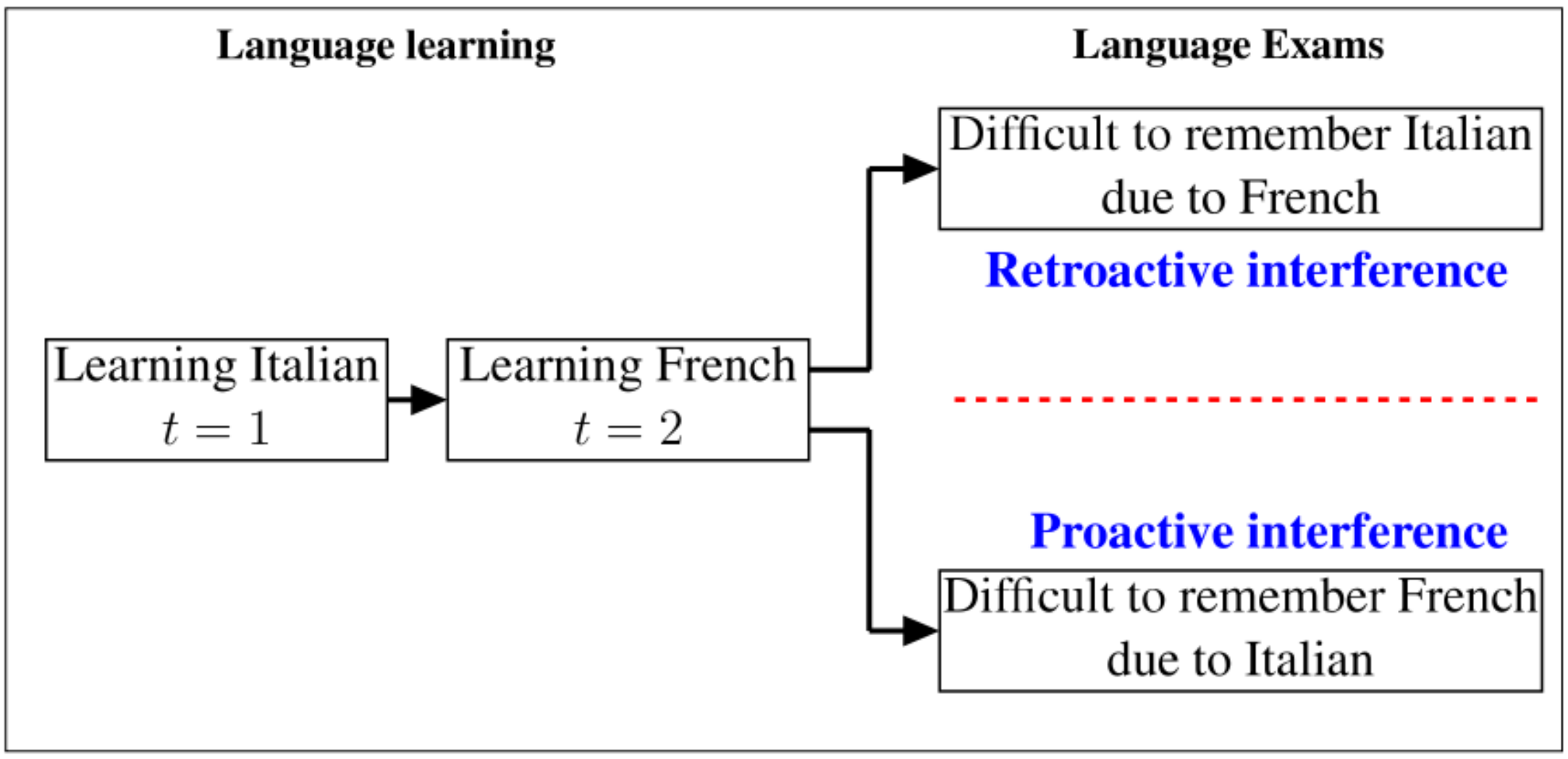

Figure 6: Example of retroactive interference and proactive interference with language learning. An individual who had learned Italian in the past ($t = 1$), has followed this with the more recent experience of learning French ($t = 2$). Retroactive interference refers to the case when the individual encounters difficulties to recall Italian words due to the more recent knowledge (French), whereas the proactive interference phenomenon explains the contrasting case when the individual tries to speak French, yet finds out that the old knowledge of Italian, which in such a case is better maintained in their memory, interferes with French.

## 6. Discussion

In this section, we aim at summarizing some of the ideas presented in this document via providing a brief discussion of the main challenges that currently encounter BCL as well as directions that, we conjecture, are potentially worth pursuing in future research.

### 6.1 Principal Challenges

We begin by discussing the most influential challenges which stand in the way of further progressing BCL.

#### 6.1.1 Catastrophic Forgetting

This remains the key challenge not only in BCL, but in the whole CL paradigm. BCL models should be stable enough to ensure that the knowledge acquired from recent tasks does not excessively overwrite the old knowledge acquired from tasks encountered in the past. Moreover, BCL models must balance the stability required for retaining old knowledge with the plasticity required to incorporate new knowledge. Throughout this manuscript, we have shed light on various mechanisms via which different BCL algorithms address catastrophic forgetting. We solely aim at summarizing this herein. Each approach to BCL aims to provide a manner based on which catastrophic forgetting can be alleviated. For example, in the regularization-based approach, forgetting is mitigated by protecting parameters which are expected to have an enormous impact on the prediction process of most previous tasks against massive changes. On the other hand, the rest of the parameters are allowed more freedom to change. Note that Bayesian CL algorithms are much more equipped than other CL algorithms to estimate which parameters are expected to be more influential than others given their innate ability to quantify uncertainty.

As mentioned earlier, exact inference is not feasible in BCL. Hence, BCL algorithms perform approximate inference which usually involves repeated approximations throughout the subsequent CL tasks. Replay-based BCL algorithms aim to alleviate the resulting risk of catastrophic forgetting via either storing or generating subsets of old data from every previous task to refresh the memory of their models. Even though this is an effective technique to address catastrophic forgetting, it entails the added burden of the storage and computational constraints resulting from storing and/or replaying old data.

The architecture-based approach provides a structured treatment to balance old and new knowledge by (rigidly) dividing the architecture into shared parts that are protected against radical changes, and task-specific parts that are allowed more freedom to change. Disadvantages of this standard form of the architecture-based approach include scalability issues potentially resulting from expanding this architecture for a large number of tasks. Moreover, this rigid division of the architecture does not take into account the heterogeneity level among the tasks. For instance, old tasks which are massively different from recent tasks can still be forgotten if the shared part of the architecture is rigidly pre-specified to be too large.

An alternative approach to address catastrophic forgetting is based on aiming to balance stability and plasticity via adjusting the learning rate in order to control the degree of adaptation that should be captured by the CL model (Pham et al., 2021). Similarly, the learning rate is adapted on a task by task basis based on the estimated degree of uncertainty in the probability distributions of the neural network weights in (Ebrahimi et al., 2020).

#### 6.1.2 TASK INTERFERENCE

Continual learning is based on transferring knowledge among tasks. Knowledge transfer can be beneficial if tasks are sufficiently similar such that the transferred knowledge becomes actually useful, which in such a case is referred to as positive transfer. Otherwise, undesirable interference can occur when knowledge is shared among dissimilar tasks, potentially leading to negative transfer. Regularization BCL methods (Kirkpatrick et al., 2017; Nguyen et al., 2018; Loo et al., 2021) can provide uncertainty estimates over which parameters should be protected against radical changes. This can play a key role in distinguishing between transferable and non-transferable knowledge. In addition, the division between task-specific parts and shared parts adopted by methods of the architecture-based approach (Adel et al., 2020; Kessler et al., 2021) can help mitigate interference. This is due to the fact that task-respective knowledge would be rather protected by the task-specific parts of the architecture, leading to reduced interference and a level of balance between stability and plasticity.

### 6.1.3 SCALABILITY

Bayesian methods normally incur a considerable computational overhead given the need to successively approximate the posterior with every upcoming task. Bayesian approaches to CL may as well struggle to scale with high-dimensional data since updating the posterior for every weight of the neural network becomes more complex. For example, in the EWC algorithm (Kirkpatrick et al., 2017), posterior updates depend on computing the Fisher information matrix, which is a very computationally intensive operation. This issue is acknowledged by the BCL community, and some variations have already been proposed as workarounds, e.g. an online version of EWC (Chaudhry et al., 2018) that is more efficient. Scalability is also strongly linked with the fact that exact Bayesian inference is often intractable for CL problems, as well as the fact that posterior approximations with higher fidelity are usually more computationally intensive than crude posterior approximations.

### 6.1.4 DEARTH OF BAYESIAN CLASS-INCREMENTAL LEARNING ALGORITHMS

As alluded to earlier, with the exception of the Bayesian FSCIL algorithm introduced in (Adel, 2025), CIL algorithms have hardly benefited from the Bayesian paradigm so far. There is a shortage of CIL algorithms which base their incremental learning process on Bayesian posterior updates. One of the main reasons why Bayesian CIL is currently understudied is the added levels of difficulty in CIL, compared to TIL. In addition to the already required approximations of the posterior in the case when task IDs are available (i.e. in TIL), the unavailability of such task IDs of every class during the test phase in CIL introduces additional challenges like the need to distinguish between all the encountered classes regardless of their respective tasks. As a result, the bulk of the aforementioned challenges, particularly computational overhead and scalability, become more cumbersome with CIL.

### 6.1.5 TASK-FREE CONTINUAL LEARNING

The vast majority of the CL algorithms in the literature (for both TIL and CIL) assume that tasks are separated via pre-defined hard boundaries (task-aware CL). Upon observing such hard boundaries, the training procedure can accordingly be divided into phases where each phase corresponds with a task. As such, hard boundaries between tasks are crucial in consolidating the learning procedure of every task. The availability of crisp boundaries represents an important basis for several key decisions in task-aware CL, like which information to conserve from previous tasks for the sake of avoiding catastrophic forgetting. Other examples of key decisions that are normally triggered based on assuming the capacity of the model to identify data belonging to each task (thanks to the hard boundaries between sequentially arriving tasks) include shuffling data within the same task since they satisfy the i.i.d. assumption conditioned on the task [3].

A more challenging, yet more realistic, CL setting that is referred to as task-free CL (Zeno et al., 2018; Aljundi et al., 2019; Lee et al., 2020; Jin et al., 2021; Pourcel et al., 2022; Wang et al., 2022; Ye & Bors, 2022b, 2023; Adel, 2024; Ye & Bors, 2025), assumes that the data distribution can change gradually and that, consequently, data should be processed in a streaming fashion without the notion of separate tasks. Thus, with gradually shifting distributions in an online setup, there are no hard boundaries between tasks since there is no available knowledge of where (data belonging to) each

3. It is true that, as previously noted, the CL paradigm assumes non i.i.d. data *overall*, which means that data belonging to different tasks are non i.i.d.. However, the i.i.d. assumption *locally* holds for data belonging to the same task in the case when tasks are separated via hard boundaries.

task would begin or end. The bulk of the Bayesian methods addressing CL depend on the availability of task IDs, at least during training. The non-availability of task boundaries, during both training and test, means that there are no available task IDs either. Hence, task-free CL depicts another challenge for the Bayesian approach to CL, especially parametric Bayesian methods.

Having said that, the work in (Lee et al., 2020) has aimed to provide a Bayesian approach to address the task-free CL setting. They adopt an architecture-based approach where the model's architecture is expanded to accommodate new data based on a mixture of experts (MoE, Jacobs et al., 1991). There is a set of experts, each of which is in charge of a subset of the data in the data stream. A Bayesian non-parametrics approach is adopted for the MoE which is a Dirichlet process mixture (DPM, Antoniak, 1974; Ferguson, 1983) model, and variational inference is used to approximate the posterior of the DPM models.

## 6.2 Potential Future Directions

### 6.2.1 Towards More Scalable Bayesian CL Algorithms

Scalability issues are very dominant in Bayesian CL. On a high level, when the number of tasks grows massively, approximating the posterior in a manner that provides a balanced treatment of the stability-plasticity tradeoff becomes gradually more difficult to achieve by BCL algorithms. BCL algorithms which rigidly pre-specify all of the critical modeling choices a priori are the most prone to the aforementioned scalability issues. To that end, adaptive BCL provides an alternative approach where some key decisions can be flexibly taken in a data-driven manner. We conjecture, also given evidence from previous works, that adaptive BCL is crucial for further improving the scalability of Bayesian CL.

Scalability challenges become even more prevalent with the architecture-based approach to BCL. With a medium-to-large number of CL tasks, expanding the architecture of the model becomes prohibitively expensive. An interesting direction to address these issues is based on adaptively identifying the proper amount of sharing between tasks in a data-driven manner. The aforementioned algorithm CLAW (Adel et al., 2020) depicts a pioneering methodology in adaptive BCL via performing automatic, data-driven adaptation of the architecture based on solely adding three parameters per neuron, and while avoiding to expand the architecture with any new neurons nor layers, potentially leading to a more scalable approach to architecture-based BCL algorithms that adaptively optimizes the tradeoff between mitigating catastrophic forgetting and improving task transfer. Extending the ideas presented in CLAW can be an intriguing direction for future research. On a high level, the adaptation ideas therein can be modified and then applied to other BCL approaches, e.g. the replay-based approach.

Another important example of how adaptive CL can mitigate the scalability problems encountering BCL is the algorithm in (Ahn et al., 2019). The memory requirements of the regularization-based BCL approach are prohibitively expensive. This includes the Fisher information matrix for the MAP estimation strategy, and the numerous variance terms required for VCL (a separate variance term for each network weight). The work in (Ahn et al., 2019) entails a method that alleviates such memory requirements via constraining all the learnable variances of the weights of the same hidden unit to have the same value. This ultimately associates the learnable variances with the hidden units of the neural network, rather than with the network weight parameters, ultimately leading to a considerable reduction in the memory requirements. We conjecture that this idea can as well be generalized to other approaches where the concept of constraining the learnable parameters should be balanced with the potential loss in the fidelity of the resulting posterior.

#### 6.2.2 BALANCING KNOWLEDGE TRANSFER

As mentioned in Section 6.1.2, knowledge transfer in CL can be either beneficial, leading to positive transfer, or harmful in which case it would lead to negative transfer. The architecture-based approach can in theory be useful herein via protecting the task-specific knowledge by the task-specific parts of the architecture. Nonetheless, achieving the right balance between task adaptation and (mitigating) task interference remains a key challenge, especially with pre-identified shared vs. task-specific components of the architecture. The (non-Bayesian) algorithm proposed in (Adel, 2024) proposes an approach addressing how to adaptively balance CL architectures based on assessing the similarity between the current task and the previously encountered tasks, while also introducing an adaptive layer that is located between the shared and task-specific components of the architecture in order to balance the tradeoff between task adaptation and mitigating task interference in a data-driven manner. Proposing Bayesian solutions based on the same idea represents a promising direction to address task interference in BCL.

#### 6.2.3 BAYESIAN CLASS-INCREMENTAL LEARNING

We have already underlined the current dearth of Bayesian CIL algorithms in the literature. If we strictly stick to our definition of Bayesian CL and CIL, the algorithm proposed in (Adel, 2025) would currently be considered the sole archetype of Bayesian CIL and Bayesian FSCIL in the literature. Notwithstanding the complexity of the challenges, the Bayesian approach can as well offer outlets for some of the outstanding issues in CIL. For example, based on uncertainty quantification, which is inherent within the Bayesian paradigm, Bayesian CIL models should be proposed to balance which information should be captured (in order to learn the new classes) along with the subsets of old information that should be conserved (to help with the retention of the previously learned classes). Basing this balanced treatment on Bayesian uncertainty estimates can mitigate overfitting which currently represents one of the most substantial issues with the bulk of CIL algorithms.

#### 6.2.4 INSPIRATION BY DEVELOPMENTAL PSYCHOLOGY

We discussed analogies between continual learning and developmental psychology in Section 5. Here, we aim to focus on some take-away aspects that can depict potentially beneficial directions for research on BCL.

Contextual adaptation was discussed n Section 5 where humans possess the innate ability to strategically prioritize either stability or flexibility given the necessities of the situation in hand. This is one area which can provide inspiration for BCL algorithms. The bulk of the current CL algorithms address the stability-plasticity tradeoff based on developing a methodology that remains fixed when encountering every upcoming task. Contextual adaptation can provide a fresh perspective where strategies addressing the stability-plasticity tradeoff are adaptable, possibly leading to a favored treatment of either stability or plasticity based on the task in hand. One way to perform strategies of the kind can be via automating the selection of key hyperparameters which control the level of stability (vs. plasticity) such that their values can be learned in a data-driven manner.

Importantly, the retrieval failure theory, which depicts one of the five popular theories of forgetting in developmental psychology, can provide another fresh perspective of addressing forgetting in BCL.

Current CL research is almost single-mindedly circulated around interpreting forgetting of CL algorithms as the outcome of completely losing the respective information. The retrieval failure theory provides a more comprehensive interpretation of forgetting where another potential cause of forgetting can be the inability to access the required information. In other words, the required information might not be completely lost, yet it cannot be retrieved. Research on BCL can benefit from this interpretation via presenting mechanisms that act as advanced retrieval cues to facilitate the retrieval of information, especially when the number of tasks grows massively.

One other aspect that can provide further inspiration for BCL is to develop informative priors that induce BCL policies promoting positive transfer, similar to what was already developed in (Schnaus et al., 2023). This can be further driven via proposing ideas similar to what an adult provides children with in scaffolding. When encountering new tasks, adults guide children in acquiring new skills that build on the knowledge they already possess. This can be seen as an efficient manner of enforcing positive transfer which is based on the adult's prior knowledge of the tasks encountered by the children. Informative priors can therefore potentially replace the role played by adults in scaffolding, such that positive transfer can be enhanced.

## 7. Conclusion

We have provided a comprehensive survey of different aspects related to the Bayesian approach to continual learning where Bayesian inference is leveraged to perform online updating over the model parameters. In Bayesian continual learning, knowledge of the continual learning model is represented as a probability distribution over its parameters that is being updated continually. As such, the old posterior continually turns into the new prior in order to allow for the integration of new knowledge while retaining old knowledge, in a manner that is more robust and adaptive compared to traditional continual learning.

After outlining definitions of continual learning, Bayesian continual learning, and the respective evaluation metrics, we have covered two different settings of the Bayesian continual learning paradigm, namely task-incremental learning and class-incremental learning. We have also examined the links between continual learning and several related paradigms such as meta-learning, multi-task learning and domain adaptation. We then moved on to presenting a taxonomy of Bayesian continual learning, along with illustrations of the most seminal algorithms in each category. Moreover, we have introduced analogies between the (Bayesian) continual learning paradigm and developmental psychology. The relationship between continual learning and developmental psychology, which has never been inspected thoroughly prior to this work, is introduced mainly so as to highlight the fact that such analogies potentially provide inspiration for future directions on Bayesian continual learning. Finally, we have discussed the most compelling challenges, and then proposed potential directions for future research on Bayesian continual learning.